\documentclass[journal]{IEEEtran}

\ifCLASSINFOpdf
   \usepackage[pdftex]{graphicx}
   \graphicspath{{../pdf/}{../jpeg/}}
   \DeclareGraphicsExtensions{.pdf,.jpeg,.png}
\else
   \usepackage[dvips]{graphicx}
   \graphicspath{{../eps/}}
   \DeclareGraphicsExtensions{.eps}
\fi
\usepackage{amsmath}
\usepackage{todonotes}
\usepackage{amsfonts}

\usepackage{graphicx} % for pdf, bitmapped graphics files
\usepackage{epsfig} % for postscript graphics files

\usepackage{booktabs}
\usepackage{siunitx}
\usepackage{soul,framed}
\usepackage{color}
\usepackage[normalem]{ulem} %show strikethrough 

\usepackage{colortbl}
\usepackage{arydshln}
\usepackage[flushleft]{threeparttable}

\definecolor{darkgreen}{rgb}{0,0.6,0.2}

\begin{document}
%
% paper title
% Titles are generally capitalized except for words such as a, an, and, as,
% at, but, by, for, in, nor, of, on, or, the, to and up, which are usually
% not capitalized unless they are the first or last word of the title.
% Linebreaks \\ can be used within to get better formatting as desired.
% Do not put math or special symbols in the title.
\title{Dense Dilated Convolutions Merging Network\\ for Land Cover Classification}
%
%
% author names and IEEE memberships
% note positions of commas and nonbreaking spaces ( ~ ) LaTeX will not break
% a structure at a ~ so this keeps an author's name from being broken across
% two lines.
% use \thanks{} to gain access to the first footnote area
% a separate \thanks must be used for each paragraph as LaTeX2e's \thanks
% was not built to handle multiple paragraphs
%

\author{Qinghui Liu$^{1,2}$,~\IEEEmembership{Student Member,~IEEE,} Michael Kampffmeyer$^{2}$,~\IEEEmembership{Member,~IEEE,} Robert Jenssen$^{2,1}$,~\IEEEmembership{Member,~IEEE,} and Arnt-B{\o}rre Salberg$^{1}$,~\IEEEmembership{Member,~IEEE}\\  % <-this % stops a space
%\thanks{*This work was supported by Norwegian Computing Center}% <-this % stops a space
\thanks{$^{1}$Norwegian Computing Center, Dept. SAMBA, P.O. Box 114 Blindern, NO-0314 OSLO, Norway}%
%\thanks{$^{1}$Q. Liu is a PhD candidate in machine learning at UiT the
%        Arctic University of Norway, Tromsø, Norway
%        }%
%\thanks{$^{2}$A.-B. Salberg is a Senior Research Scientist with the Norwegian Computing Center, Oslo, Norway.
%        }%
\thanks{$^{2}$UiT Machine Learning Group, Department of Physics and Technology, UiT the Arctic University of Norway, Troms{\o}, Norway
        }%
}

% note the % following the last \IEEEmembership and also \thanks - 
% these prevent an unwanted space from occurring between the last author name
% and the end of the author line. i.e., if you had this:
% 
% \author{....lastname \thanks{...} \thanks{...} }
%                     ^------------^------------^----Do not want these spaces!
%
% a space would be appended to the last name and could cause every name on that
% line to be shifted left slightly. This is one of those "LaTeX things". For
% instance, "\textbf{A} \textbf{B}" will typeset as "A B" not "AB". To get
% "AB" then you have to do: "\textbf{A}\textbf{B}"
% \thanks is no different in this regard, so shield the last } of each \thanks
% that ends a line with a % and do not let a space in before the next \thanks.
% Spaces after \IEEEmembership other than the last one are OK (and needed) as
% you are supposed to have spaces between the names. For what it is worth,
% this is a minor point as most people would not even notice if the said evil
% space somehow managed to creep in.

% The paper headers
\markboth{IEEE Transactions on Geoscience and Remote Sensing}%
{Shell \MakeLowercase{\textit{et al.}}: Dense Dilated Convolutions Merging Network}
% The only time the second header will appear is for the odd numbered pages
% after the title page when using the twoside option.
% 
% *** Note that you probably will NOT want to include the author's ***
% *** name in the headers of peer review papers.                   ***
% You can use \ifCLASSOPTIONpeerreview for conditional compilation here if
% you desire.

% If you want to put a publisher's ID mark on the page you can do it like
% this:
%\IEEEpubid{0000--0000/00\$00.00~\copyright~2015 IEEE}
% Remember, if you use this you must call \IEEEpubidadjcol in the second
% column for its text to clear the IEEEpubid mark.

% use for special paper notices
%\IEEEspecialpapernotice{(Invited Paper)}

% make the title area
\maketitle

% As a general rule, do not put math, special symbols or citations
% in the abstract or keywords.
\begin{abstract}
Land cover classification of remote sensing images is a challenging task due to limited amounts of annotated data, highly imbalanced classes, frequent incorrect pixel-level annotations, and an inherent complexity in the semantic segmentation task.
% , which presents significant domain gap from existing state-of-the-art segmentation models in ground imageries. 
% \commentM{The previous sentence is a bit unclear. Do you mean that there is a domain gap between land cover classification and the more commonly encountered images in CV (e.g. hand-held camera)?} \commentAB{I agree with Michael. Maybe we should just delete the the part "...which presents signficant ....imageries.} 
In this work, we propose a novel architecture called the Dense Dilated Convolutions Merging Network (DDCM-Net) to address this task. The proposed DDCM-Net consists of dense dilated image convolutions merged with varying dilation rates. This effectively utilizes rich combinations of dilated convolutions that enlarge the network's receptive fields with less parameters and features compared to the state-of-the-art approaches in the remote sensing domain.
% This effectively enlarges the kernels' receptive fields using less parameters compared to the current state-of-the-art. 
Importantly, DDCM-Net obtains fused local and global context information, in effect incorporating surrounding discriminative capability for multi-scale and complex shaped objects with similar color and textures in very high resolution aerial imagery.
% \commentAB{Can we justify the claim "rotational scale-invariant representations"?} 
We demonstrate the effectiveness, robustness and flexibility of the proposed DDCM-Net on the publicly available ISPRS Potsdam and Vaihingen data, as well as the DeepGlobe land cover dataset. Our single model, trained on 3-band Potsdam and Vaihingen data, achieves better accuracy in terms of both mean intersection over union (mIoU) and F1-score compared to other published models trained with more than 3-band data. We further validate our model on the DeepGlobe dataset, achieving state-of-the-art result 56.2\% mIoU with much less parameters and at a lower computational cost compared to related recent work.
% \commentAB{Could you specify the mIoU numbers and F1-score values you achieved?}
% \commentAB{What state-of-the-art performance are you referring to for the DeepGlobe dataset?}
% We further adapted our model for DeepGlobe domain, which also shows state-of-the-art performance with much less parameters and computational cost. 
% \commentM{We further validate our model on the DeepGlobe dataset, achieving state-of-the-art performance with much less parameters and computational cost.}

\end{abstract}

% Note that keywords are not normally used for peerreview papers.
\begin{IEEEkeywords}
Deep learning, very high resolution (VHR) optical imagery, land cover classification, semantic segmentation
\end{IEEEkeywords}

% For peer review papers, you can put extra information on the cover
% page as needed:
% \ifCLASSOPTIONpeerreview
% \begin{center} \bfseries EDICS Category: 3-BBND \end{center}
% \fi
%
% For peerreview papers, this IEEEtran command inserts a page break and
% creates the second title. It will be ignored for other modes.
\IEEEpeerreviewmaketitle

\section{Introduction}
% The very first letter is a 2 line initial drop letter followed
% by the rest of the first word in caps.
% 
% form to use if the first word consists of a single letter:
% \IEEEPARstart{A}{demo} file is ....
% 
% form to use if you need the single drop letter followed by
% normal text (unknown if ever used by the IEEE):
% \IEEEPARstart{A}{}demo file is ....
% 
% Some journals put the first two words in caps:
% \IEEEPARstart{T}{his demo} file is ....
% 
% Here we have the typical use of a "T" for an initial drop letter
% and "HIS" in caps to complete the first word.
\IEEEPARstart{A}{utomatic} semantic classification of land cover in remote sensing data is of great importance for sustainable development, autonomous agriculture, and urban planning. Thanks to the progress achieved in the deep learning and computer vision community on natural images, most deep learning architectures \cite{long2015fully, ronneberger2015u, badrinarayanan2017segnet, zhao2016pyramid, wang2017understanding, peng2017large} for semantic segmentation can also be used for land cover classification tasks in the remote sensing domain. Semantic segmentation refers to the assignment of a semantic category to every pixel in the images, which in this work consist of very high resolution (VHR) aerial images. Currently, the state-of-the-art end-to-end semantic segmentation models are mostly inspired by the idea of fully convolutional networks (FCNs), which generally consist of an encoder-decoder architecture \cite{long2015fully}. All layers in the encoder and decoder modules are based on convolutional neural networks (CNN). 
% \commentR{Link CNNs and FCNs. Reader may not know the link.} 
However, to achieve higher performance, FCN-based end-to-end methods normally rely on deep and wide multi-scale CNN architectures that typically require a large number of trainable parameters and computation resources. In addition, there is also a lot of redundancy in deep CNNs that often results in vanishing gradients in backward propagation, diminishing feature reuse in forward propagation, and long training time \cite{huang2016deep}. 

 \begin{figure}[htpb!]
 \centering
  \includegraphics[width=0.48\textwidth]{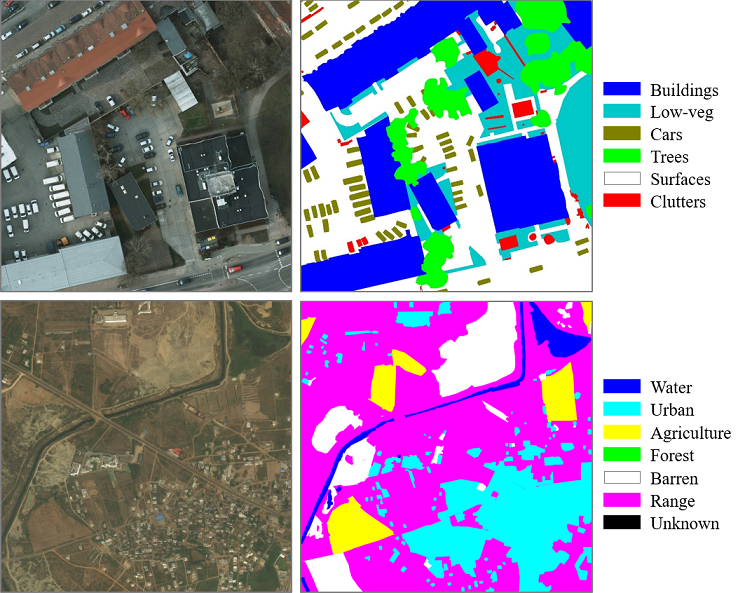} %,height=8.5cm]{testallpng.png}
      \caption{Examples of land cover labels (right) and corresponding remote sensing images (left) from two different datasets (ISPRS Potsdam \cite{ISPRS2018} and DeepGlobel \cite{DeepGlobe18}) separately. Semantic label colors are shown beside the ground truths.}
  \label{fig:semapping}
\end{figure}

In general, VHR remote sensing images contain diverse objects and intricate variations in their aspect-ratio and color-textures (e.g. roads, roofs, shadows of buildings, low plants and branches of trees \cite{ISPRS2018}). Moreover, many remote sensing images are completely consisting of "stuff" classes (amorphous regions such as forest, agricultural areas, water and so on). This brings challenges for semantic mapping in remote sensing images. Fig.~\ref{fig:semapping} shows illustrative examples of land cover classification for remote sensing images. Thus, richer and multi-scale global contextual representations play a key role in land cover mapping for VHR aerial images. 

In this work, we propose a novel network architecture, called the dense dilated convolutions merging network (DDCM-Net), which utilizes multiple dilated convolutions with various dilation rates. The proposed network learns with densely linked dilated convolutions and outputs a fusion of all intermediate features without losing resolutions during the extraction of multi-scale features. This significantly reduces the  computational redundancies and costs. Our experiments demonstrate that the network achieves robust and accurate results on the representative ISPRS 2D semantic labeling datasets \cite{ISPRS2018}. Motivated by the recent success of depthwise separable convolutions \cite{chen2018deeplab}, we also explore grouped convolutions \cite{pelt2018mixed} with strided operations adapted into our DDCM-Net, which is shown to further improve net speed and accuracy. We finally demonstrate the effectiveness of the adapted DDCM-Net on the DeepGlobe land cover challenge dataset \cite{DeepGlobe18}.
% and attain a performance of 56.2\% on the hold-out validation set, setting a new state-of-the-art. 
% \commentAB{This is not the place to state results. Make sure that the performance is specified in the results. The 56.2\% should also be stated in the abstract. By the way: what kind of performance is this?}
In summary, our contributions are: 
\begin{enumerate}
  \item We propose a new computationally light-weight and scalable architecture based on dilated convolutions \cite{yu2015multi} that can be used as a simple, yet effective, encoder or decoder module for semantic segmentation tasks.
  \item In the proposed network, one can arbitrarily control the depths, widths, groups and strides of the modules with various dilation rates in order to address different problems. 
  \item Our proposed end-to-end model outperforms or achieves competitive performance on different representative remote sensing datasets compared to other published related methods.
\end{enumerate}

A preliminary version of this paper appeared in \cite{liu_2019}. Here, we extend our work by (i) extending the methodology with two variants where group and strided convolutions are exploited to further boost the model's flexibility;
(ii) expanding the experiment section by including more public datasets, providing more detailed training details and presenting additional result comparisons;
(iii) providing in-depth discussions in terms of model's dilation and density policies, generalization, as well as a more detailed analysis of computation complexity;
(iv) providing a more thorough review of related work.

The paper is structured as follows. Section~\ref{backg} provides an overview of the related work. Section~\ref{data} introduces the datasets used in our work. In Section~\ref{method}, we present the methodology in details. Experimental procedure and evaluation of the proposed method is performed in Section~\ref{exp}. Section~\ref{disc} provides discussion of our results, and, finally in Section~\ref{concl}, we draw conclusions.

\section{Background}
\label{backg}
Deep learning and CNNs have been revolutionary for computer vision and image classification \cite{krizhevsky2012imagenet, farabet2013learning} in particular. Even though segmentation can be viewed as a pixel-to-pixel classification problem, most modern CNN models for semantic segmentation are inspired by fully convolutional networks (FCNs) \cite{long2015fully}. The FCN was the first CNN model without any fully connected layers that was trained in an end-to-end manner directly classifying each pixel to its corresponding label. However, vanilla FCNs generally cause loss of spatial information due to the presence of pooling layers that reduce the resolutions of feature maps by sacrificing the positional information of objects. In order to alleviate this issue, U-Net \cite{ronneberger2015u} extends the FCN by introducing skip connections between the encoder and decoder modules. In the decoder module, the spatial information is gradually recovered by fusing skipped connections with upsampling layers or de-convolution layers. Since then, the encoder-decoder architecture has been widely extended in recent works including SegNet \cite{audebert2017segment}, GCN \cite{peng2017large}, PSPNet \cite{zhao2016pyramid}, DUC \cite{wang2017understanding}, DeepLabV3+ \cite{chen2018deeplab} and so on. In general, these architectures differ from each other in how they capture rich and global contextual information at multiple scales. For instance, PSPNet \cite{zhao2016pyramid} introduces a pyramid pooling module to aggregate the context by applying large kernel pooling layers, while DeepLabV3+ \cite{chen2018deeplab} utilizes several parallel atrous convolution with different rates (called Atrous Spatial Pyramid Pooling). Similarly, the authors of \cite{2019sdc} presented a unified descriptor network for dense matching tasks, so called SDC-stacked dilated convolution, which combines parallel dilated convolutions with different dilation rates of ($[1, 2, 3, 4]$). Instead of the parallel combination methods, a cascading structure of dilated convolution layers was first presented in \cite{yu2015multi} with exponentially increasing rates of dilation that achieved state-of-the-art results on a natural image segmentation benchmarks in that year. The authors of \cite{strubell2017fast} have also proposed a sequential structure of iterating dilated convolutions which demonstrated higher accuracy with impressive speed improvements in contrast to the previously best performing model BI-LSTM-CRF \cite{huang2015bidirectional}, for the sequence labeling tasks when processing entire documents at a time.

In contrast, our novel architecture has three major differences. Firstly, we sequentially stack the output of each layer with its input features before feeding it to the next layer in order to alleviate context information loss. Secondly, our final output is computed on all features generated by intermediate layers, which can effectively aggregate the fused receptive field of each layer and maximally utilize multi-scale context information. Thirdly, our method is much more flexible and extendable with group and strided convolutions to address different domain problems.

Our applied focus in this paper is land cover classification based on remote sensing. Lately, the  FCNs and encoder-decoder architectures have been widely adapted and applied to the ISPRS \cite{ISPRS2018} Semantic Labeling Contest \cite{ paisitkriangkrai2015effective, Sherrah16, lin2016efficient, MarmanisSWGDS16, audebert2016semantic, audebert2017joint, wang2017gated, MouRiFCN2018, michael2018, liu2019semantic}, and the DeepGlobe CVPR-2018 \cite{DeepGlobe18} challenge of automatic classification of land cover types \cite{samy2018nu, davydow2018land, pascual2018uncertainty, seferbekov2018feature, ghosh2018stacked, kuo2018deep,tian2018dense}. Paisitkriangkrai et al. \cite{paisitkriangkrai2015effective} proposed a scheme for high-resolution land cover classification using a combination of a patch-based CNN and a random forest classification that is trained on hand-crafted features. Further, Sherrah \cite{Sherrah16} applied FCNs to semantic labelling of aerial imagery and illustrated that higher accuracy can be achieved than with more traditional patch-based approaches.
% to have achieved higher accuracy than the patch-based approach.
% \commentM{to have achieved higher accuracy than the patch-based approach -- and illustrated that higher accuracy can be achieved than with more traditional patch-based approaches.}
In addition, deep learning has also been exploited for multi-modal data processing in remote sensing. For instance, Audebert et al. \cite{audebert2017joint} proposed a multi-scale SegNet approach (so-called FuseNet) to leverage both a large spatial context and the high resolution data, while early and late fusion strategies of multi-modality data are also exploited. However, such fusion techniques require all modalities to be available to the classification during both training and testing. The authors of \cite{michael2018} therefore presented a novel CNN architecture based on so-called hallucination networks for urban land cover classification, able to replace missing data modalities in the test phase. This enables fusion capabilities even when data modalities are missing in testing.

Recently, the authors of \cite{MouRiFCN2018} proposed a recurrent network in fully convolutional network (RiFCN) trying to better fuse multi-level features with boundary-aware features to achieve fine-gained inferences. Similarly, the stacked U-Nets architecture is proposed in \cite{ghosh2018stacked} for ground material segmentation in remote sensing imagery, which merges high-resolution details and the long distance context information captured at low-resolution to generate segmentation maps. Further, Kuo et al. \cite{kuo2018deep} introduced an aggregation decoder in combination with DeepLabV3 architecture to fuse different-level features progressively from the encoder for final prediction, while the authors of \cite{tian2018dense} proposed a dense fusion classmate network (DFCNet) which tried to fuse auxiliary training data as "classmate" to capture supplementary features for land cover classification. One of the main ideas behind all the architectures is to take into account the multi-level context to improve the prediction of the segmentation. Even though these state-of-the-art designs could alleviate the loss of global contextual information, they are often computationally expensive with a lot of redundancy in order to capture dense and multi-scale contextual features \cite{huang2016deep}. In the following, we demonstrate that the DDCM-Net achieves competitive results or outperforms for land cover classification on benchmark datasets at a lower computational cost compared to related recent work.

\section{Benchmark Datasets}
\label{data}
In this paper, we focus on two publicly used databases, namely the ISPRS 2D semantic labeling contest datasets \cite{ISPRS2018}, and the DeepGlobe land cover challenge dataset \cite{DeepGlobe18}. 
The ISPRS datasets are comprised of aerial images over two cities in Germany: Potsdam\footnote{http://www2.isprs.org/commissions/comm3/wg4/2d-sem-label-potsdam.html} and Vaihingen\footnote{http://www2.isprs.org/commissions/comm3/wg4/2d-sem-label-vaihingen.html}, which have been labelled with six of the most common land cover classes: impervious surfaces, buildings, low vegetation, trees, cars and clutter. The DeepGlobe land cover dataset consists of satellite data collected from the DigitalGlobe Vivid+ dataset \cite{DeepGlobe18}, and focuses on rural areas. This includes seven types of land covers: urban (man-made, built up areas with human artifacts), agriculture (farms, cropland, orchards, vineyards, ornamental horticultural areas, and so on), rangeland (any non-forest, non-farm, green land and grass), forest (any land with at least 20\% tree crown density plus clear cuts), water (rivers, oceans, lakes, wetland, ponds), barren (mountain, rock, dessert, beach, land with no vegetation), and unknown (clounds and others). Each dataset provides online leaderboards and reports test metrics measured on hold-out test images. 

\subsubsection{Potsdam}
The Potsdam dataset consists of 38 tiles of size $6000 \times 6000$ pixels with a ground resolution of 5cm. 14 of these are used as hold-out test images. Tiles consist of Red-Green-Blue-Infrared (RGB-IR) four-channel images. While both the digital surface model (DSM) and normalized DSM (nDSM) data are also included in the dataset, we only focus on the 3-channel RGB images in this work. 

\subsubsection{Vaihingen}
The Vaihingen dataset contains 33 tiles of varying size (on average approximately $2100 \times 2100$ pixels) with a ground resolution of 9cm, of which 17 are used as hold-out test images. Tiles are composed of Infrared-Red-Green (IRRG) 3-channel images. Though DSMs and nDSMs data are also available for all images in the dataset, we only use IRRG data in this paper. 

\subsubsection{DeepGlobe}
DeepGlobe Land Cover data contains $1146$ RGB images of size $2448 \times 2448$ pixels with a ground resolution of 50cm. 803 of these images have a publicly available ground truth and are used as the training set, while the remaining images are split into a hold-out validation and test set consisting of 171 and 172 images, respectively. 
% \commentM{Made some changes in the previous sentence and have a question to make sure that I did not change the meaning. Both the validation and test dataset gt are not available, right?} 
Due to the variety of land cover types and density of annotations \cite{DeepGlobe18}, this dataset is more challenging than the two above-mentioned datasets. 

\section{Methods}
\label{method}
We first briefly revisit the concept of dilated convolutions. We then present our proposed DDCM architecture, based on such dilated convolutions. Furthermore, we provide detailed information regarding the procedure for training the network.

\begin{figure*}[htbp]
 \centering
  \includegraphics[width=0.7\textwidth]{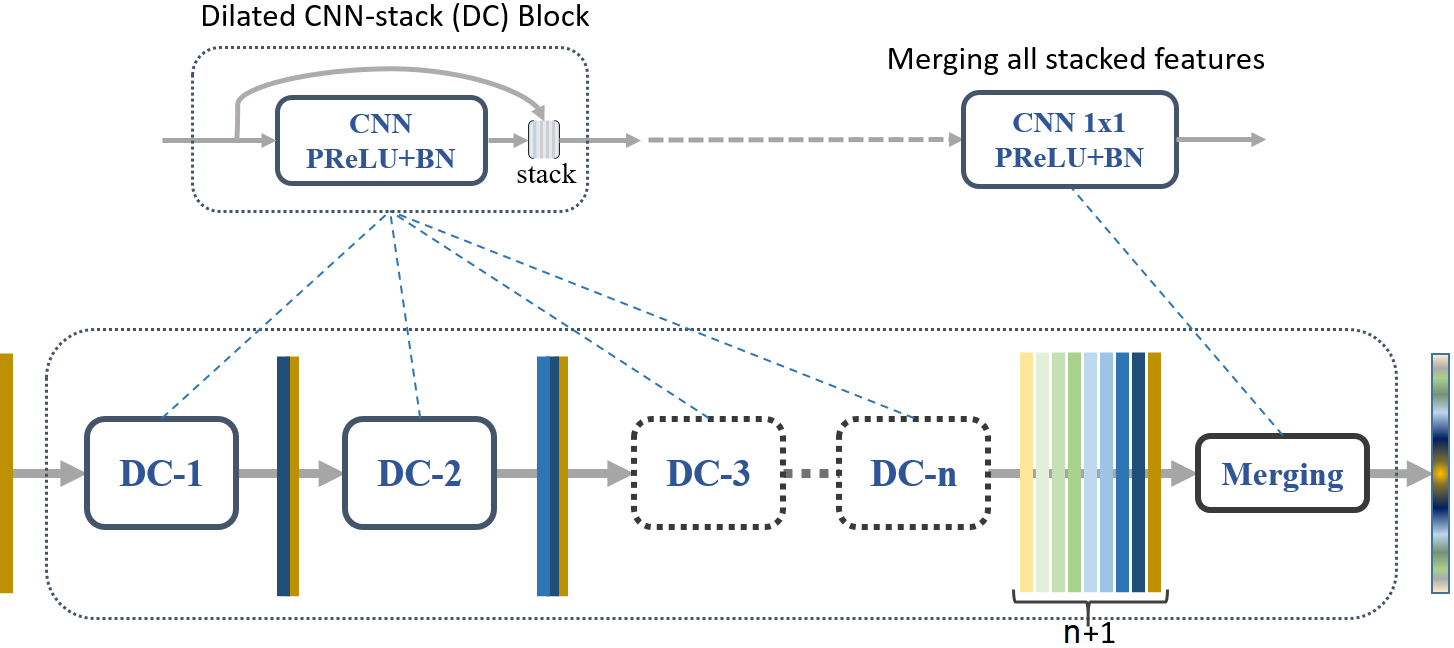} %,height=8.5cm]{testallpng.png}
  \caption{Example of the DDCM architecture composed of $n$ $\{1, 2, 3, ... , n\}$ DC blocks with various dilation rates.}
  \label{fig:ddcm}
\end{figure*}

\subsection{Dilated Convolutions}
Dilated convolutions~\cite{yu2015multi} have been demonstrated to improve performance in many classification and segmentation tasks \cite{chen2018deeplab, pelt2018mixed, li2018csrnet, wei2018revisiting}. 
One key advantage is that they allow us to flexibly adjust the filter's receptive field to capture multi-scale information without resorting to down-scaling and up-scaling operations. A 2D dilated convolution operator can be defined as

\begin{equation}\label{dconv} 
%   X^{i,j}_{l} = 
   g(x_{\ell}) = \sum_{c \in C_{\ell}} \theta^{c}_{k, r}\ast x_{\ell}^{c}
\end{equation}
where, $\ast$ denotes a convolution operator, $g:\mathbb{R}^{H_{\ell}\times W_{\ell}\times C_{\ell}} \rightarrow\mathbb{R}^{H_{\ell+1}\times W_{\ell+l}}$ convolves the input feature map $x_{\ell}\in \mathbb{R}^{H_{\ell}\times W_{\ell}\times  C_{\ell}}$. A dilated convolution $\theta_{k, r}$ with a filter size $k$ and dilation rate $r \in \mathbb{Z}^+$ is only nonzero for a multiple of $r$ pixels from the center. In a dilated convolution, a kernel size $k$ is effectively enlarged to $k+(k-1)(r-1)$ with the dilation factor $r$. As a special case, a dilated convolution with dilation rate $r=1$ corresponds to a standard convolution. 

% \begin{equation}\label{dconv} 
% %   X^{i,j}_{l} = 
%   g_{i,j}(x_{\ell}) = \sum_{c=0}^{C_{\ell}} \theta^{c,i,j}_{k, r}\ast x_{\ell}^{c}
% \end{equation}
% where, $\ast$ denotes a convolution operator, $g_{i,j}:\mathbb{R}^{H_{\ell}\times W_{\ell}\times C_{\ell}} \rightarrow\mathbb{R}^{H_{\ell+1}\times W_{\ell+l}}$ convolves the input feature map $x_{\ell}\in \mathbb{R}^{H_{\ell}\times W_{\ell}\times  C_{\ell}}$ within channel $c \in \{0,1,\dotsc, C_{\ell}\}$ at row $i$ and column $j$. A dilated convolution $\theta^{c}_{k, r}$ with a filter size $k$ and dilation rate $r \in \mathbb{Z}^+$ is only nonzero for a multiple of $r$ pixels from the center. In dilated convolution, a kernel size $k$ is effectively enlarged to $k+(k-1)(r-1)$ with the dilation factor $r$. As a special case, a dilated convolution with dilation rate $r=1$ corresponds to a standard convolution. 

\subsection{Dense Dilated Convolutions Merging Module}

 \begin{figure}[htbp]
 \centering
  \includegraphics[width=0.45\textwidth]{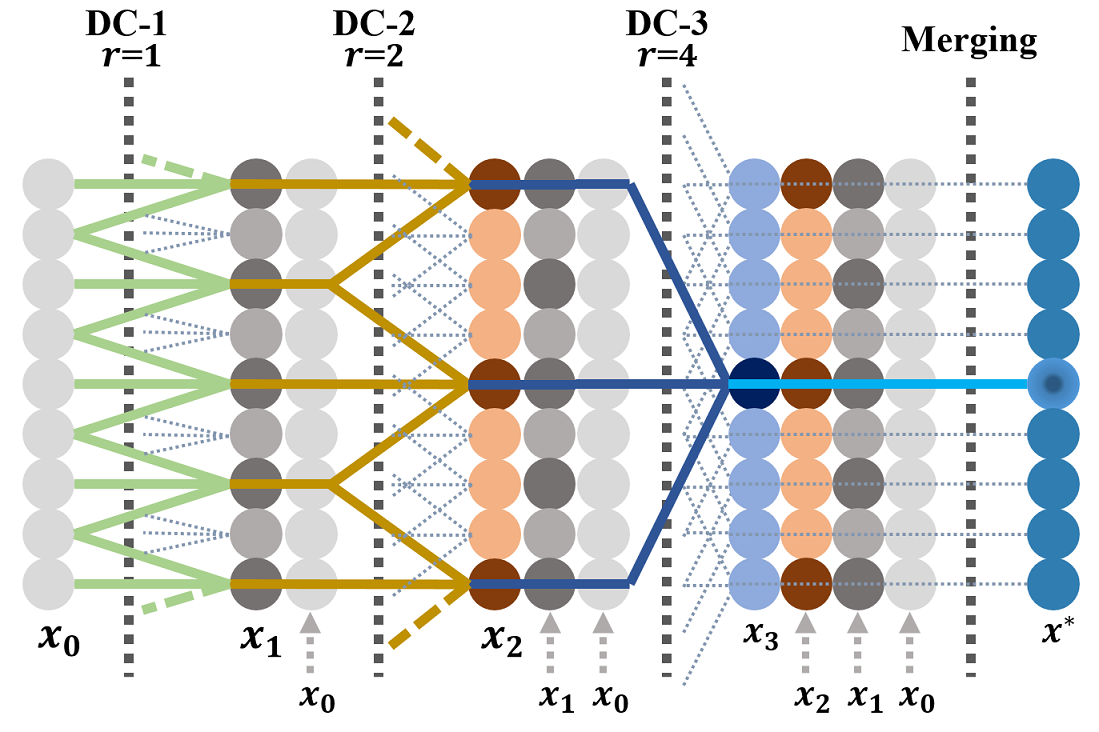} %,height=8.5cm]{testallpng.png}
  \caption{A simple 1-D example of the DDCM module composed of three DC blocks (kernel size = 3) with dilation rates of $1$, $2$ and $4$. Here we can see that $x_1$ is produced from $x_0$ by a 1-dilated convolution with a receptive field of 3. $x_2$ is produced from $[x_1, x_0]$ by a 2-dilated convolution with fused receptive fields of $[7, 5]$.  $x_3$ is produced from $[x_2, x_1, x_0]$ by a 4-dilated convolution with fused receptive fields of $[15, 11, 9]$. The final output $x^{\ast}$ is produced from $[x_3, x_2, x_1, x_0]$ by the so-called merging layer, which fuses multi-scale context information by aggregating various receptive fields of $[15, 11, 9, 7, 5, 3, 1]$.}
  \label{fig:dcs}
\end{figure}

The Dense Dilated Convolutions Merging Module (DDCM-Module) consists of a number of Dilated CNN-stack (DC) blocks with a merging module as output as shown in Fig.~\ref{fig:ddcm}. A basic DC block is composed of a dilated convolution followed by PReLU \cite{he2015delving} non-linear activation and batch normalization (BN) \cite{ioffe2015batch}. It then stacks the output with its input and feeds the stacked data to the next layer.
% \sout{, which can alleviate loss of context information and problems with vanishing gradients when adding more layers}
% \commentAB{suggest we move the frase ", which can alleviate loss of context information and problems with vanishing gradients when adding more layers." to the discussion."}
The final network output is computed by a merging layer composed of $1\times1$ filters with PReLU and BN in order to efficiently combine all stacked features generated by intermediate DC blocks. In practice, densely connected DC blocks, typically configured with linearly or exponentially increasing dilation factors, enable DDCM networks to have very large receptive fields with just a few layers. Please note that we apply zero padding to the input of every DC block in order to keep the resolution of its output equal to the resolution of the input. 
% \commentAB{"capture scale-invariant features by merging multi-scale features properly" is just a claim. We should not say it here, maybe in the discussions, if we can justify it.}

Fig.~\ref{fig:dcs} illustrates a simple 1-D example of the DDCM module composed of three DC blocks (kernel size equal 3) with dilation rates of $[1, 2, 4]$. In the DC-1 layer, $x_1$ is produced from $x_0$ by a 1-dilated convolution, where each element of $x_1$ has a receptive field of 3. In the DC-2 layer, $x_2$ is produced from $[x_1, x_0]$ by a 2-dilated convolution. Note, the receptive field for the elements of $x_2$ are $[7,5]$. Similarly, in the DC-3 layer, $x_3$ is produced from $[x_2, x_1, x_0]$ by a 4-dilated convolution with fused receptive fields of $[15, 11, 9]$. The final output $x^{\ast}$ is thus produced from $[x_3, x_2, x_1, x_0]$ by the so-called merging layer, which fuses multi-scale context information by aggregating various receptive fields of $[15, 11, 9, 7, 5, 3, 1]$. It is easy to see that the number of parameters associated with each DC layer grows linearly, while the fused receptive field size is nearly exponentially increasing.
% \commentM{while each element in $x_1$  has fused receptive fields of $[7, 5]$ ... Should this be $x_2$? Maybe rewrite to "Note, the receptive field for the elements of $x_2$ are $[7,5]$.}

% \commentAB{The equations below need to be motivated, and Fig. 3 need to be explained better".} 
% \begin{equation}\label{dcs} 
%   x_{n} = BN_{\gamma, \beta}(P_{\alpha}(g([x_{n-1}, x_{n-2}, ..., x_0])))
% \end{equation}

% \begin{equation}\label{merge} 
%   x^{\ast} = M_{\alpha, \gamma, \beta}([x_{n}, x_{n-1}, ..., x_0], W_{1\times1\times (N+1)})
% \end{equation}
% where, $BN$ denotes mini Batch Normalization with two learnable parameters $\gamma$ and $\beta$. $P$ is PReLU function with $\alpha$ as a learnable parameter \cite{he2015delving}, and $Merging$ denotes the merging layer.
% \commentM{Might need to define some additional quantities such as x and W?} \commentAB{Agree. I have used W on the previous page, so we need a new letter.}

\begin{figure*}[htpb!]
 \centering
  \includegraphics[width=0.78\textwidth]{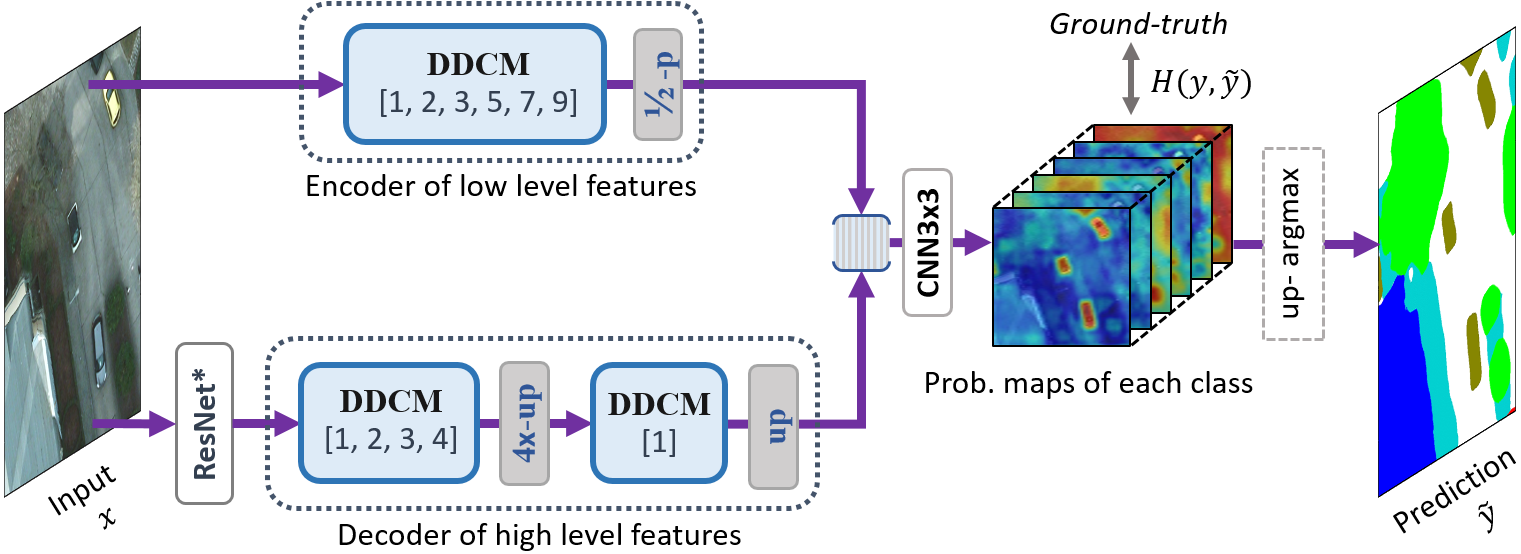} %,height=8.5cm]{testallpng.png}
  \caption{End-to.end pipeline of DDCM-Net for semantic mapping of VHR Potsdam images. The encoder of low level features encodes multi-scale contextual information from the initial input images by a DDCM module (output 3-channel) using $3\times3$ kernels with 6 different dilation rates $[1, 2, 3, 5, 7, 9]$. The decoder of high level features decodes highly abstract representations learned from a ResNet-based backbone (output 1024-channel) by 2 DDCM modules with rates $[1, 2, 3, 4]$ (output 36-channel) and $[1]$ (output 18-channel) separately. The transformed low-level and high-level feature maps by DDCMs are then fused together to infer pixel-wise class probabilities. Here, 'p' and 'up' denote pooling and up-sampling respectively.}
  \label{fig:ddcm_resnet}
\end{figure*}

\subsection{Variants of the DDCM module}
\subsubsection{Grouped convolutions}
Inspired by ResNeXt \cite{xie2017aggregated}, grouped convolutions \cite{krizhevsky2012imagenet} are also exploited in the DC blocks in order to further reduce the depth and parameter size of DDCM-net, especially when the DDCM modules are used as the decoders of high-level features. ResNeXt has demonstrated that increasing cardinality (group number) is a more effective way of gaining accuracy than going deeper or wider \cite{xie2017aggregated}. We therefore introduce a variant of DDCM, i.e. $\text{DDCM}(g=2)$, where “$g=2$” denotes the fact that grouped convolutions with 2 groups are used in the DC blocks.
% \commentM{suggests -- denotes the fact that} 

\subsubsection{Strided convolutions}
To further reduce the computational cost when increasing dilation rates, we can apply a dilated convolution with a stride of greater than one pixel, which samples only every $s$ pixels in each direction in the output. Here $s$ denotes the stride of this dilated convolution. This is similar to a two-step approach with unit stride convolution followed by downsampling, but reduces computational cost. When using a strided dilated convolution in a DC block, we therefore need to apply bilinear upsampling to scale the output to the same resolution as the input before concatenation. There are three variants of the DDCM-Module evaluated in this work, i.e. $\text{DDCM}(s=2)$, $\text{DDCM}(s=3)$ and $\text{DDCM}(s=r+1)$ with different striding strategies: a stride of 2, a stride of 3 and a dynamic stride of $(r+1)$ separately, where $r$ denotes the corresponding dilation rate.

\subsection{The DDCM network} \label{archit}
DDCM modules define building blocks from which a more complex network can be built. This is illustrated in Fig.~\ref{fig:ddcm_resnet} 
% \commentR{OR PERHAPS: DDCM modules define building blocks from which a more complex network can be built. This is illustrated in ... showing ...}
showing the end-to-end pipeline of the DDCM-Net combined with a pre-trained model for land cover classification. Compared to other encoder-decoder architectures, our proposed DDCM-Net only fuses low-level features one time before the final prediction CNN layers, instead of aggregating multi-scale features captured at many different encoder layers \cite{ronneberger2015u, peng2017large, MouRiFCN2018, davydow2018land, pascual2018uncertainty, seferbekov2018feature, ghosh2018stacked, kuo2018deep,tian2018dense}. This makes our model simple and neat, yet effective with lower computational cost. In particular, this model is easy to adapt by adjusting the density (number of the output feature maps) and dilation strategy of the encoder and/or decoder features to tackle different tasks, depending on different domains. 
% \commentAB{Do you really need to talk about the dataset here? The reviewer will then wonder; why only Potsdam? What about Vahingen and DeepGlobe?}

In our work, we only utilize the first three bottleneck layers of pretrained ResNet-based \cite{he2016deep} backbones (both ResNet50 \cite{he2016deep} and SE-ResNeXt50 \cite{seresnet}) and remove the last bottleneck layer and the fully connected layers to reduce the number of parameters to train. Furthermore, due to the larger complexity and variety of the DeepGlobe dataset compared to the ISPRS data, we utilize a $\text{DDCM}(s=2)$ module configured with larger dilation growing rates $[1, 2, 4, 8, 16, 32]$ as the low-level encoder, and two $\text{DDCM}(g=2, s=2)$ modules configured by $[1, 2, 4]$ and $[1]$ as the high-level decoder. This configuration results in feature maps of size 64-channel and 32-channel, rather than 36-channel and 18-channel for the model on ISPRS data. We also choose SE-ResNeXt50 as the backbone, instead of ResNet50.
% \commentM{maybe replace "with flexibly" with "by" and maybe remove "part" after "low level encoder" and "high level decoder"} \commentM{Suggestion, replace: ", which output more features of 64-dimension and 32-dimension separately, rather than 32-dim and 18-dim for the model on ISPRS data." with "This configuration results in feature maps of size 64-dim and 32-dim, rather than 36-dim and 18-dim for the model on ISPRS data."}
% \commentM{Question, should it be 16-dim instead of 18? Also, here you refer to the the spatial dimension, right? In that case it might not be clear that it is $64\times 64$ and not just 64 dim.}

% \commentAB{Why two branches in Fig. 4, one with low level feature and one with high level features? Is this the only way to design or can you think of building other networks using the DDCM module?}

\section{Experiments and results}
\label{exp}
In this section, we investigate the proposed network on the Potsdam (Section~\ref{exp-pots}), Vaihingen (Section~\ref{exp-pots}) and DeepGlobe (Section~\ref{exp-dg}) datasets and report both qualitative and quantitative results of multi-class land cover classification.

\subsection{Training details}
% \commentR{Question - Possible to say something like: According to best practices, we ... }
% For all experiments, 
According to best practices, we train using Adam \cite{KingmaB14adam} with AMSGrad \cite{amsgrad2018} as the optimizer with weight decay $2 \times 10^{-5}$ applied to all learnable parameters except biases and batch-norm parameters, and polynomial learning rate (LR) decay $(1 - \frac{cur\_iter}{max \_iter})^{0.9}$ with the maximum iterations of $10^8$. We also set $2 \times LR$ to all bias parameters in contrast to weights parameters. We use initial LRs of $\frac{8.5 \times 10^{-5}}{\sqrt{2}} $ and $\frac{8.5 \times 10^{-4}}{\sqrt{2}} $ for the ISPRS data and DeepGlobe data, respectively. For the training on ISPRS data (both Potsdam and Vaihingen), we utilized a stepwise LR schedule method that reduces the LR by a factor of 0.85 every 15 epochs based on our training observations and empirical evaluations, while for the training on DeepGlobe data, we utilized multi-step LR policy, which reduces the LR by a factor of 0.56 at epochs $[4, 8, 16, 24, 32, 96, 128]$ guided by our empirical results.

We apply a cross-entropy loss function with median frequency balancing (MFB) weights as defined in the equations \ref{eq:mfb} and \ref{eq:loss} \cite{kampffmeyer2016semantic}.
\begin{equation}\label{eq:mfb}
W_c =\frac{\text{median} (\{f_c | c\in \mathcal{C} \})}{f_c}, 
\end{equation}

\begin{equation}\label{eq:loss}
Loss = -\frac{1}{N} \sum\limits_{i=1}^N \sum\limits_{c=1}^C {l_c}^{(n)} \log{({p_c}^{(n)})} W_c
\end{equation}
where $W_c$ is the weight for class $c$, $f_c$ the pixel- frequency of class $c$, ${p_c}^{(n)}$ is the probability of sample belonging to class $c$, ${l_c}^{(n)}$ denotes the class label of sample $n$ in class $c$. 

We train and validate the networks for the Potsdam and Vaihingen datasets with ramdomly sampled 5000 patches of size $256\times 256$ as input and batch size of five. For the experiments on the DeepGlobe dataset, we use two down-sampled resolutions of $1224 \times 1224$ and $816 \times 816$ (down-scaled from $2448 \times 2448$), then we train with $4000$ crops ($765 \times 765$) and batch size of four. The training data is sampled uniformly and randomly shuffled for each epoch. We conduct all experiments in this paper using PyTorch \cite{NEURIPS2019_9015} on a single computer with one NVIDIA 1080Ti GPU. 
% \commentR{Possible to cite some other papers that have used similar setups?}

\subsection{Augmentation and evaluation methods}
During training on Potsdam and Vaihingen data, we randomly flip or mirror images for data augmentation (with probability $0.5$), while on the DeepGlobe data, we also augment the crops by randomly shifting (limit 0.0625), scaling (limit 0.1) and rotating them (limit 10). The albumentations library \cite{Albumentations2018} for data augmentation is utilized in this work. Please note that all training images are normalized to [0.0, 1.0] after data augmentation.

We apply test time augmentation (TTA) in terms of flipping and mirroring. For the Potsdam and Vaihingen data, we use sliding windows (with $448 \times 448$ size at a 100-pixel stride) on a test image and stitch the results together by averaging the predictions of the over-lapping TTA regions to form the output. While for DeepGlobe data, we first apply TTA on a down-sampled ($3$x) test image ($816 \times 816$) and then up-sample all the predictions back to original sizes and average them to get the final output. The performance is measured by both the F1-score \cite{kampffmeyer2016semantic}, and the mean Intersection over Union (IoU) \cite{liuqinghui2018}. Please note that the mIoU metric was computed by averaging over the six classes (excluding the 'Unknown' class) in the DeepGlobe contest. 

\subsection{Potsdam and Vaihingen} \label{exp-pots}
For evaluation, the labeled part of the Potsdam dataset is split into a training set (19 images), a validation set (2 images of 4\_10 and 7\_10), and a local test set (3 images of areas 5\_11, 6\_9 and 7\_11). The Vaihingen dataset is similarly divided into training (11 images), validation (2 images of areas 7 and 9) and local test set (4 images of areas 5, 15, 21 and 30). While the hold-out test sets contain 14 images (areas: 2\_13, 2\_14, 3\_13, 3\_14, 4\_13, 4\_14, 4\_15, 5\_13, 5\_14, 5\_15, 6\_13, 6\_14, 6\_15 and 7\_13) and 17 images (areas: 2, 4, 6, 8, 10, 12, 14, 16, 20, 22, 24, 27, 29, 31, 33, 35 and 38) for the Potsdam and Vaihingen datasets, respectively.
% \commentM{Might need to clarify local vs. hold-out test set} 
Table \ref{tab:rgbresults} shows our results on the hold-out test sets and our local test sets of ISPRS Potsdam and Vaihingen separately with a single trained model. The mean F1-score (mF1) and the mean IoU (mIou) are computed as the average measure of all classes except the clutter class. 

\begin{table}[hptb!]
\centering 
  \caption{Results on the hold-out test images of ISPRS Potsdam and Vaihingen datasets with a single trained DDCM-R50 model separately.}
\resizebox{\columnwidth}{!}{
\begin{threeparttable}
\begin{tabular}{c|p{8mm}|p{8mm}p{8mm}p{11mm}p{9mm}p{9mm}|p{9mm}} %\hline%\toprule 
    \textbf{F1-score} & $\textbf{OA}$ & \textbf{Surface} & \textbf{Building} & \textbf{Low-veg} & \textbf{Tree} & \textbf{Car}  & \textbf{mF1} \\  \hline  %\hline%\midrule 
%     &&&\\[-1em]
    Potsdam &\text{0.908} \newline0.931$^{*}$ & 0.929 \newline 0.946$^{*}$ & 0.969 \newline 0.983$^{*}$ & 0.877 \newline 0.865$^{*}$ & 0.894 \newline 0.892$^{*}$ & 0.949 \newline 0.939$^{*}$ & 0.923 \newline 0.925$^{*}$\\ \hdashline
    Vaihingen & 0.904 \newline 0.921$^{*}$ & 0.927 \newline 0.934$^{*}$ & 0.953 \newline 0.973$^{*}$ & 0.833 \newline 0.814$^{*}$ &0.894 \newline 0.914$^{*}$ &  0.883 \newline 0.909$^{*}$ & 0.898 \newline 0.909$^{*}$ \\
  \bottomrule
    \textbf{IoU} & $\textbf{}$ & \textbf{} & \textbf{} & \textbf{} & \textbf{} & \textbf{}  & \textbf{mIoU} \\  \hline  %\hline%\midrule 
%     &&&\\[-1em]
    Potsdam & 0.908 \newline0.931$^{*}$ & 0.867 \newline 0.898$^{*}$ & 0.940 \newline 0.966$^{*}$ & 0.781 \newline 0.762$^{*}$ & 0.809 \newline 0.805$^{*}$ & 0.902 \newline 0.885$^{*}$ & 0.860 \newline 0.863$^{*}$ \\
     \hdashline
    Vaihingen & 0.904 \newline 0.921$^{*}$ & 0.863 \newline 0.876$^{*}$ & 0.909 \newline 0.948$^{*}$ & 0.713 \newline 0.686$^{*}$ & 0.808 \newline 0.842$^{*}$ & 0.790 \newline 0.832$^{*}$ & 0.817 \newline 0.837$^{*}$  \\  \hline
\end{tabular}
\begin{tablenotes}
            \item[*] Results marked with $^{*}$ were measured on our local test images, others were measured on hold-out test sets (14 images and 17 images for the Potsdam and Vaihingen separately).
    \end{tablenotes}
  \end{threeparttable}
} 
\label{tab:rgbresults}%

\end{table}

We also compare our results to other related published work on the ISPRS Potsdam RGB dataset and Vaihingen IRRG dataset. These results are shown in Table~\ref{tab:potsdam_scores} and \ref{tab:vaihingen_scores} respectively. Our single model with overall F1-score (92.3\%) on Potsdam RGB dataset, achieves around 0.5 percent higher score compared to the second best model - FuseNet+OSM \cite{audebert2017joint}.
% \commentAB{The "In other words..." sentence is discussion. Suggest you move it to that chapter.} 
Similarly, our model trained on Vaihingen IRRG images, also obtained the best overall performance with 89.8\% F1-score that is around 1.1\% higher than the score of the second best model - GSN \cite{wang2017gated}. It is also worth noting that, although our OA is only marginal better ($+0.1$\%) for Vaihingen, and even worse ($-1.5$\%) for Potsdam, our model obtained better F1 scores. We therefore believe that our proposed method has better capability to handle extremely unbalanced classes. Further, by balancing and modelling the surrounding classes (such as road/surface and buildings) more accurately with our model, the car class will be easier to distinguish and thus has better results with increased receptive fields as shown in Table~\ref{tab:vaihingen_scores}.

\begin{figure*}[htpb!]
 \centering
  \includegraphics[width=0.98\textwidth]{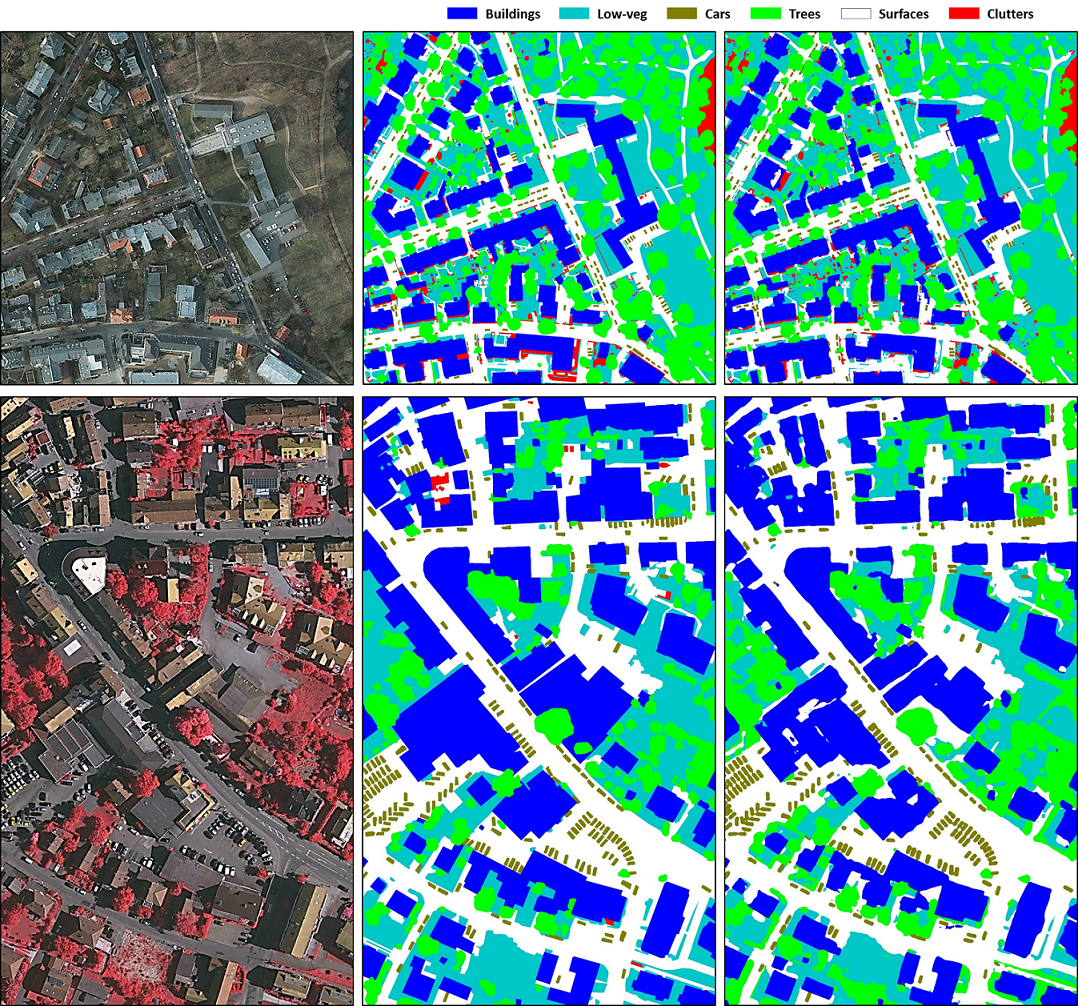} %,height=8.5cm]{testallpng.png}
  \caption{Mapping results for test images of Potsdam tile-3\_14 (top) and Vaihingen tile-27 (bottom). From the left to right, the input images (left), the ground truths (middle) and the predictions of our single DDCM-R50 model.}
  \label{fig:test}
\end{figure*}

Fig.~\ref{fig:test} shows the qualitative comparisons of the land cover mapping results from our model and the ground truths on the test set. We observe that our model is able to segment both large multi-scale objects (such as buildings) and small objects (such as cars) very well with fine-gained boundary recovery without any post-processing.

\begin{table}[hptb!]
\centering 
  \caption{Comparisons between our method with other published methods on the hold-out RGB test images of ISPRS Potsdam dataset.}
\resizebox{\columnwidth}{!}{
\begin{threeparttable}
\begin{tabular}{c|p{9mm}|p{9mm}p{8mm}p{11mm}p{9mm}p{9mm}|p{8mm}} \hline%\toprule
    \textbf{Models} & $\textbf{OA}$ & \textbf{Surface} &\textbf{Building} & \textbf{Low-veg} & \textbf{Tree} &  \textbf{Car} & \textbf{mF1} \\  \hline  \hline%\midrule 
%     &&&\\[-1em] 
    HED+SEG.H-Sc1 \cite{MarmanisSWGDS16} & 0.851  & 0.850  & 0.967  & 0.842  & 0.686  & 0.858  & 0.846 \\  
    RiFCN \cite{MouRiFCN2018}            & 0.883  & 0.917  & 0.930  & 0.837  & 0.819   & 0.937  & 0.861 \\
    RGB+I-ensemble \cite{michael2018}    & 0.900  & 0.870  & 0.936  & 0.822  & 0.845  & 0.892 & 0.873 \\
    Hallucination \cite{michael2018}     & 0.901  & 0.873  & 0.938  & 0.821   & 0.848  & 0.882  & 0.872 \\
    DNN\_HCRF \cite{liu2019semantic}     & 0.884  & 0.912  & 0.946  & 0.851 & 0.851 & 0.928  & 0.898\\ 
    SegNet RGB \cite{audebert2017joint}  & 0.897  & 0.930  & 0.929  & 0.850  & 0.851  & 0.951  & 0.902\\ 
    DST\_2 \cite{Sherrah16}              & 0.903  & 0.925  & 0.964  & 0.867 & 0.880  & 0.947 & 0.917 \\
    FuseNet+OSM \cite{audebert2017joint} & \cellcolor{gray!25} \textbf{0.923}  & \cellcolor{gray!25}\textbf{0.953}  & 0.959  & 0.863 & 0.851  & \cellcolor{gray!25}\textbf{0.968}  & 0.918\\  \hline 
    \textbf{Ours} & $\textbf{}$ & \textbf{} & \textbf{} & \textbf{} & \textbf{} & \textbf{}  & {} \\  \hline
    DDCM-R50 & 0.908 \newline (-1.5\%) & 0.929\newline (-2.4\%)  & \cellcolor{gray!25}\textbf{0.969} \newline (+0.5\%)   & \cellcolor{gray!25}\textbf{0.877} \newline (+1.0\%)  &\cellcolor{gray!25}\textbf{0.894} \newline (+1.4\%)  & 0.949\newline (-1.9\%)  & \cellcolor{gray!25}\textbf{0.923} \newline (+0.5\%)\\ \hdashline % 
     $\text{DDCM}(s=2)$ & 0.908  & 0.930  & 0.968  & 0.876 & 0.895  & \textbf{0.952}  & \textbf{0.924} \\ 
    $\text{DDCM}(s=3)$ & 0.910  & 0.932  & 0.967  & 0.878 & 0.895  & 0.937 & 0.922\\ 
    $\text{DDCM}(s=r+1)$ & 0.911  & 0.933  & 0.968  & 0.876 & 0.894  & 0.950  & 0.924 \\ \hline
\end{tabular}
% \begin{tablenotes}
%             \item[*] Evaluation on eroded boundary ground truths of tiles ID 5\_11, 6\_9, 7\_11, .
% %             \item[+] OSM means OpenStreetMap data.
%         \end{tablenotes}
  \end{threeparttable}
} 
\label{tab:potsdam_scores}%

\end{table}

\begin{table}[hptb!]
\centering 
  \caption{Comparisons between our method with other published methods on the hold-out IRRG test images of ISPRS Vaihingen Dataset.} 
  
\resizebox{\columnwidth}{!}{
\begin{threeparttable}
\begin{tabular}{c|p{9mm}|p{9mm}p{8mm}p{11mm}p{9mm}p{9mm}|p{8mm}} \hline%\toprule
    \textbf{Models} & $\textbf{OA}$ & \textbf{Surface} & \textbf{Building} & \textbf{Low-veg} & \textbf{Tree} & \textbf{Car}  & \textbf{mF1} \\  \hline \hline%\midrule 
%     &&&\\[-1em] 
    UOA \cite{lin2016efficient}                 & 0.876  & 0.898  & 0.921  & 0.804 & 0.882  & 0.820  & 0.865 \\  
    DNN\_HCRF \cite{liu2019semantic}            & 0.878  & 0.901  & 0.932  & 0.814 & 0.872  & 0.720  & 0.848\\ 
    ADL\_3 \cite{paisitkriangkrai2015effective} & 0.880  & 0.895  & 0.932  & 0.823 & 0.882  & 0.633  & 0.833 \\ 
    DST\_2 \cite{Sherrah16}                     & 0.891  & 0.905  & 0.937  & 0.834 & 0.892  & 0.726 & 0.859 \\ 
    ONE\_7 \cite{audebert2016semantic}          & 0.898  & 0.910  & 0.945  & \cellcolor{gray!25}\textbf{0.844} & 0.899  & 0.778 & 0.875\\ 
    DLR\_9 \cite{MarmanisSWGDS16}               & 0.903  & 0.924  & 0.952  & 0.839 & 0.899  & 0.812  & 0.885 \\  
    GSN \cite{wang2017gated}                    & 0.903  & 0.922  & 0.951  & 0.837 & \cellcolor{gray!25}\textbf{0.899} & 0.824  & 0.887 \\ \hline 
    \textbf{Ours} & $\textbf{}$ & \textbf{} & \textbf{} & \textbf{} & \textbf{} & \textbf{}  & {} \\  \hline
    DDCM-R50  & \cellcolor{gray!25}\textbf{0.904} \newline (+0.1\%) & \cellcolor{gray!25}\textbf{0.927}\newline (+0.3\%)& \cellcolor{gray!25}\textbf{0.953} \newline (+0.1\%)   & 0.833 \newline (-1.1\%)   &0.894 \newline (-0.5\%)  & \cellcolor{gray!25}\textbf{0.883}\newline (+5.9\%) & \cellcolor{gray!25}\textbf{0.898} \newline (+1.1\%) \\ \hdashline % 
     $\text{DDCM}(s=2)$ & 0.901  & 0.924  & 0.951  & 0.826 & 0.891  & \textbf{0.890}  & 0.896 \\ 
    $\text{DDCM}(s=3)$ & 0.901  & 0.923  & 0.951  & 0.828 & 0.891  & 0.879 & 0.894\\ 
    $\text{DDCM}(s={r+1})$ & 0.901  & 0.923  & 0.949  & 0.829 & 0.892  & 0.888  & 0.896 \\ \hline
\end{tabular}
% \begin{tablenotes}
%             \item[*] \hl{Evaluated on all test IRRG images of tiles ID 2, 4, 6, 8, 10, 12, 14, 16, 20, 22, 24, 27, 29, 31, 33, 35 and 38 with only IRRG bands.} 
%         \end{tablenotes}
  \end{threeparttable}
} 
\label{tab:vaihingen_scores}%

\end{table}

\begin{table}[hptb!]
\centering 
  \caption{Performance comparisons on the hold-out validation set of DeepGlobe data with other published methods.}
\resizebox{0.75\columnwidth}{!}{
\begin{threeparttable}
\begin{tabular}{c|p{15mm}c} \hline %\toprule 
\textbf{Models}  &  \textbf{mIoU} & \textbf{GFLOPs} \\ \hline  \hline%\midrule 
NU-Net \cite{samy2018nu}  & 0.428 & - \\
InceptionV3+Haralick \cite{davydow2018land} & 0.476 & -  \\  
GCN-based \cite{pascual2018uncertainty}  & 0.485 & - \\
FPN \cite{seferbekov2018feature} & 0.493 &-  \\
Stacked U-Nets \cite{ghosh2018stacked} & 0.507& -  \\
DeepLabv3+ \cite{kuo2018deep} & 0.510 &-  \\
ClassmateNet \cite{tian2018dense} & 0.519 & -  \\ 
DFCNet \cite{tian2018dense} & 0.526 & -  \\ 
Deep Aggregation Net \cite{kuo2018deep} & 0.527 &- \\ \hline
\textbf{Ours} & $\textbf{}$ & \textbf{} \\  \hline
$\text{DDCM-SER50}$  &\cellcolor{gray!25} \textbf{0.562} & \cellcolor{gray!25}\textbf{4.68}  \\ \hline
%\bottomrule
\end{tabular}
% \begin{tablenotes}
%     \item[*] Our model is composed of $\text{DDCM}^{c2}_{s^2}$ modules with the first 3 bottleneck layers of SE-ResNeXt50 \cite{seresnet} as the backbone.
% \end{tablenotes}
\end{threeparttable}
} 
\label{tab:dg_results}%
% $\text{DDCM}^{c2}_{s^2}\text{-SER50}$
\end{table}

\subsection{DeepGlobe}\label{exp-dg}
\begin{figure}[htpb!]
 \centering
  \includegraphics[width=0.48\textwidth]{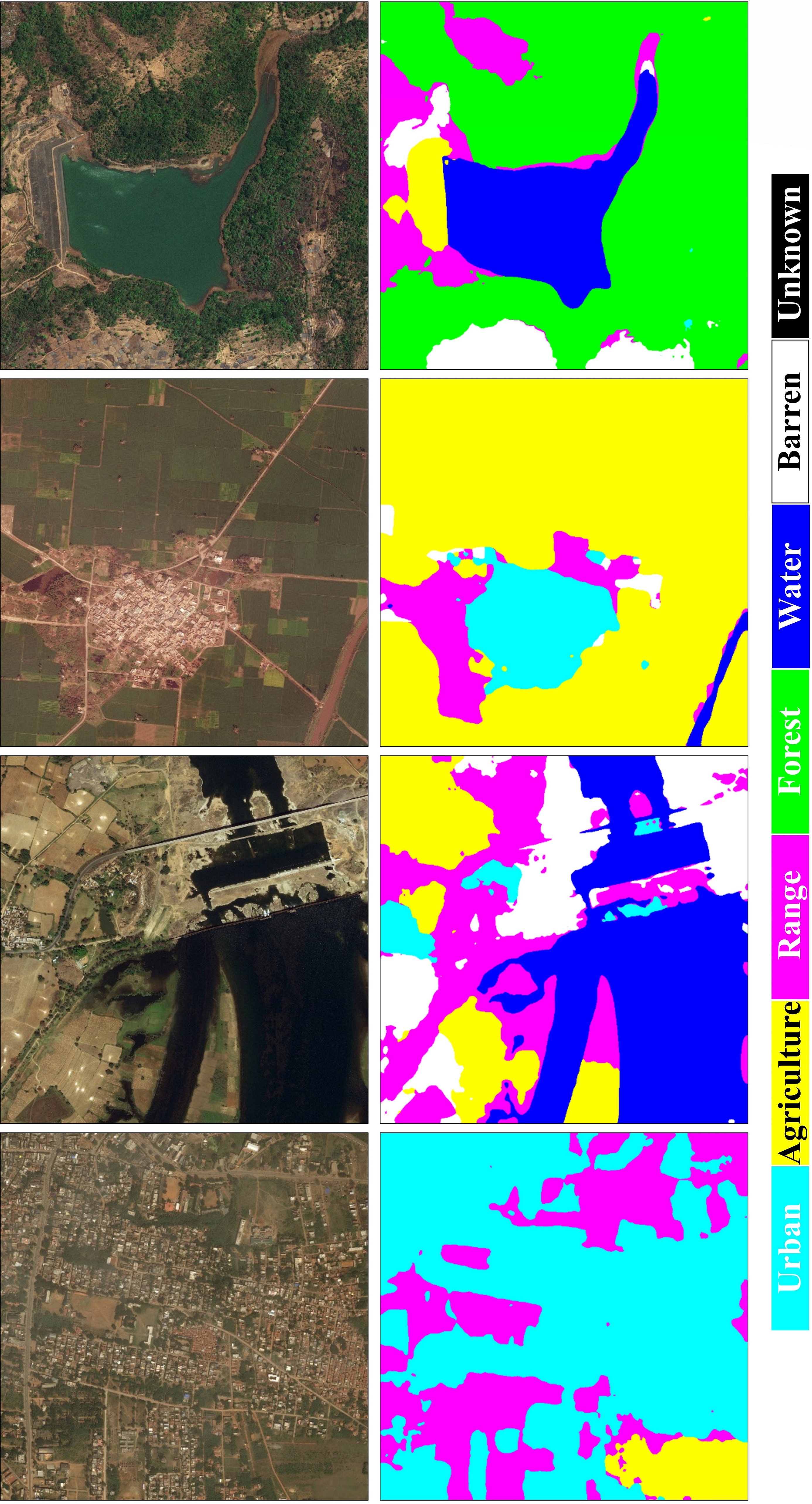} %,height=8.5cm]{testallpng.png}
  \caption{Mapping results on hold-out validation images of DeepGlobe. From the left to right, the input satellite images and the predictions of our model.}
  \label{fig:db_test_real}
\end{figure}

% \commentR{Suggest to rethink this sentence. "on the leaderboard" is not precise, nor is "our models"}
Our DDCM-SER50 model achieves new state-of-the-art result with 56.2\% mIoU on DeepGlobe land cover classification challenge dataset. As shown in Table \ref{tab:dg_results}, we compare our DDCM network with other published models (\cite{samy2018nu, davydow2018land, pascual2018uncertainty, seferbekov2018feature, ghosh2018stacked, kuo2018deep, tian2018dense}) on the hold-out validation set (the public leaderboard \footnote{https://competitions.codalab.org/competitions/18468\#results} up to the date of May 1, 2019). Our model obtained above 3.5\% higher mIoU than the second best model \cite{kuo2018deep}. Fig.~\ref{fig:db_test_real} shows some examples of the predictions on the hold-out validation images. We didn't apply any post-processing (e.g. graph-based fine tuning as utilized in \cite{kuo2018deep}), nor any multi-scale prediction fusion methods as applied in \cite{tian2018dense}. We just make the prediction on the $3\times$ down scaling of the test image and then perform up-sampling back to its original resolution. 

\begin{table}[hptb!]
\centering 
  \caption{Quantitative Comparison of parameters size, FLOPs (measured on input image size of $3 \times 256 \times 256$), Inference time on CPU and GPU separately, and mIoU  on ISPRS Potsdam RGB dataset.}
\resizebox{0.95\columnwidth}{!}{
\begin{threeparttable}
\begin{tabular}{c|c|p{13mm}p{11mm}p{21mm}p{11mm}} \hline %\toprule 
\textbf{Models} & \textbf{Backbones} & \textbf{Parameters} \newline (Million)& \textbf{FLOPs} \newline (Giga) & \textbf{Inference time} \newline (ms - CPU/GPU) &\textbf{mIoU}$^{*}$  \\ \hline  \hline%\midrule 
U-Net \cite{ronneberger2015u}& VGG16 & 31.04 & 15.25 & 1460 / \textbf{6.37}  & 0.715 \\
FCN8s \cite{long2015fully} & VGG16 & 134.30 & 73.46 & 6353 / 20.68 & 0.728 \\
SegNet \cite{badrinarayanan2017segnet}& VGG19 & 39.79 & 60.88 & 5757 / 15.47 & 0.781 \\
GCN \cite{peng2017large} & ResNet50 & 23.84 & 5.61  & 593 / 11.93 & 0.774 \\
PSPNet \cite{zhao2016pyramid} & ResNet50 & 46.59 & 44.40 & 2881 / 81.08 & 0.789 \\
DUC \cite{wang2017understanding} & ResNet50 & 30.59 & 32.26 & 2086 / 68.24 & 0.793 \\  \hline 
\textbf{Ours} & $\textbf{}$ & \textbf{} & \textbf{} & \textbf{} \\  \hline
DDCM-R50 & $\text{ResNet50}^\dagger$ & \cellcolor{gray!25}\textbf{9.99}  & \cellcolor{gray!25}\textbf{4.86} & \cellcolor{gray!25}\textbf{238} / 10.23 &\cellcolor{gray!25} \textbf{0.808} \\ \hdashline
$\text{DDCM}(s=2)$ & $\text{ResNet50}^\dagger$ & 9.99  &4.48 & {159} / 11.39 &\textbf{0.811}     \\
$\text{DDCM}(s=3)$ & $\text{ResNet50}^\dagger$ & 9.99  &4.43 & {144} / 11.25 &0.798     \\
$\text{DDCM}(s=r+1)$& $\text{ResNet50}^\dagger$ & 9.99  &4.42 & \textbf{132} / 11.50 &0.810  \\ 
% \hdashline
% $\text{DDCM}^{c2}_{s^2}\text{-SER50}$ & $\text{SE-ResNeXt50}^\dagger$ & 11.37  &4.68 &\textbf{-}     \\
\hline
%\bottomrule
\end{tabular}
\begin{tablenotes}
    \item[*] mIoU was measured on full reference ground truths of our local test images 5\_11, 6\_9 and 7\_11 in order to fairly compare with our previous work \cite{liuqinghui2018}.
    \item[*] Inference time was measured on CPU - AMD Ryzen Threadripper 1950X and GPU - NVIDIA GeForce GTX 1080Ti respectively. 
\end{tablenotes}
\end{threeparttable}
} 
\label{tab:parameters}%

\end{table}

\begin{table}[hptb!]
\centering 
  \caption{Ablation studies for our proposed method on the hold-out RGB test images of the ISPRS Potsdam dataset.}
\resizebox{\columnwidth}{!}{
\begin{threeparttable}
\begin{tabular}{c|p{9mm}|p{9mm}p{9mm}p{11mm}p{9mm}p{9mm}|p{9mm}} \hline%\toprule
    \textbf{Models} & $\textbf{OA}$ & \textbf{Surface} &\textbf{Building} & \textbf{Low-veg} & \textbf{Tree} &  \textbf{Car} & \textbf{mF1} \\  \hline  \hline%\midrule 
%     &&&\\[-1em] 
    DDCM-R50 & \textbf{0.908}  & \textbf{0.929} & \textbf{0.969} & \textbf{0.877}  & \textbf{0.894}  & \textbf{0.949} & \textbf{0.923} \\ \hdashline % 
     \text{No-LL-Encoder} & 0.899 \newline (-0.9\%)  & \cellcolor{gray!25}\text{0.919} \newline (-1.0\%)  & \cellcolor{gray!25}\text{0.948} \newline (-2.1\%) & 0.869 \newline (-0.8\%) & 0.893 \newline (-0.1\%) & \cellcolor{gray!25}\text{0.936} \newline (-1.3\%) & 0.913 \newline (-1.0\%) \\ \hdashline
    \text{No-Dilation} & 0.892 \newline (-1.6\%)  & \cellcolor{gray!25}\text{0.910} \newline (-1.9\%) & \cellcolor{gray!25}\text{0.938} \newline (-3.1\%) & 0.868 \newline (-0.9\%)& 0.892 \newline (-0.2\%)  & \cellcolor{gray!25}\text{0.929} \newline (-2.0\%)  & 0.908 \newline (-1.5\%) \\ \hline
\end{tabular}
\begin{tablenotes}
            \item[*] No-LL-Encoder means the model was configured without the low-level encoder stream, while adjusting the output channels of the high level decoder to 21 instead of 18 in the standard DDCM-R50 model. The reason for increasing the number of the high level channels is to counter the loss of features from the low-level encoder stream.
            \item[*] No-Dilation means that each DDCM module in the DDCM-R50 model only used unit dilation rates for its convolution layers.
        \end{tablenotes}
  \end{threeparttable}
} 
\label{tab:no_encoder_scores}%

\end{table}

\section{Discussion} \label{disc}
\subsection{Preliminary analysis}
As a baseline, we re-implemented, trained and evaluated some popular architectures on the local Potsdam test set \cite{liuqinghui2018}. 
We compared our methods to them in terms of parameters sizes, computational cost (FLOPs), inference time on both CPU (AMD Ryzen Threadripper 1950X) and GPU (NVIDIA GeForce GTX 1080Ti), and mIoU evaluated on the full reference ground truths of the dataset. Table~\ref{tab:parameters} details the quantitative results of our DDCM-R50 model against others. Our model consumes about 9x and 13x less FLOPs with 4x and 4.7x fewer parameters and 12x and 24x faster inference speed on CPU, but achieves +1.9\% and +2.7\% higher mIoU than PSPNet \cite{zhao2016pyramid} and SegNet \cite{badrinarayanan2017segnet} respectively.

Additionally, we also investigated the effectiveness of strided convolutions with the purpose of reducing the computational cost of dilated modules. We observe that with a dynamic stride of $r+1$, our model has the best speed without loss of the F1-scores in comparison to a stride of 2 or stride-1 model as shown the final hold-out tests as shown in Table~\ref{tab:potsdam_scores}. And interestingly, we find that both dynamic strided and stride-2 policies could improve the models accuracy of small objects (i.e cars). Overall, $\text{DDCM}{(s=2)}$ based mode obtained the best performance on small objects (cars) with $+0.3$\% and $+0.7$\% higher IoU than standard convolutions on both the Potsdam and Vaihingen test datasets, respectively, as shown in Tables~\ref{tab:potsdam_scores} and \ref{tab:vaihingen_scores}. We therefore believe that dilated convolutions with strided operations (i.e a stride of 2) could not only improve a model's computational efficiency, but also capture better contextual representations that further boost the model's capability for detailed object boundary recovery.

\begin{figure*}[htpb!]
 \centering
  \includegraphics[width=0.98\textwidth]{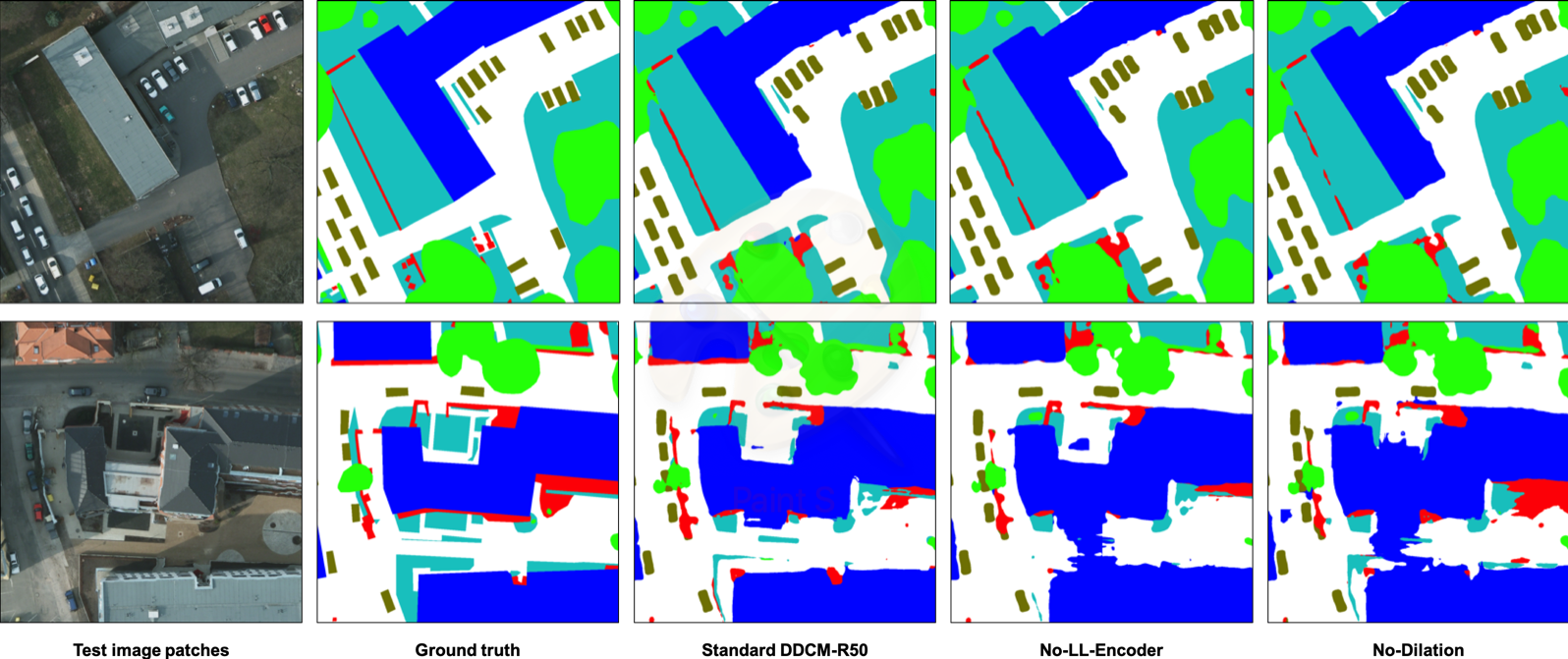} %,height=8.5cm]{testallpng.png}
  \caption{Mapping results comparison. From the left to right, the input image patches, the ground truth (left), the standard DDCM-R50, the no-low-level-encoder DDCM-R50, and the no-dilation DDCM-R50 (right).}
  \label{fig:no_dilation_compare}
\end{figure*}

\subsection{Flexible dilation and density policies}
We used two different dilation settings strategies. For ISPRS data, we configured a DDCM module with linearly growing dilation rates ($[1, 2, 3, 5, 7, 9]$) as the low-level features encoder, while for the DeepGlobe dataset, we build the encoder with exponentially growing dilation rates ($[1, 2, 4, 8, 16, 32]$) with a stride of two, since we see that the DeepGlobe images contain more spatially chaotic objects with lower resolutions, larger scales and less geometrical attributes than the ISPRS images. We believe DDCM modules with bigger dilation configurations could capture larger multi-scale and global context in this case, but require more computational cost. Hence, one has to make some trade-offs on the dilation policies and the densities based both on the dataset and on the budget of computation resources. Similarly, as for the decoders of high-level features, we also follow different strategies in terms of the dilation settings and densities of DDCM modules for the ISPRS and DeepGlobe data as described in Section~\ref{archit}. In particular, we adopt both strided and grouped convolutions together that use stride equal to two and group equal to two in the DDCM modules with the output densities of 64-channel and 32-channel respectively. These settings strike the best trade-off between speed and accuracy on the DeepGlobe database in our experiments.

We also evaluated the the influence of the low-level feature encoder and the dilated strategies. We performed two ablation studies, by training the following two models: 1) The No-LL-Encoder model that was configured without the low-level encoder stream, but only used the high level decoder branch which output 21 channels instead of 18 channels in the standard DDCM-R50 model; 2) The No-Dilation model that only uses standard convolutions by fixing the dilation rate $r=1$ for each of the convolutional layers in the DDCM-R50 model. Table~\ref{tab:no_encoder_scores} shows the final test results on the Potsdam dataset. The performance, in terms of mF1, dropped off overall 1.0\% and 1.5\% with No-LL-Encoder and No-Dilation, respectively. Performance is particularly decreasing for the building (-2.1\% and -3.1\%) and car (-1.3\% and -2.0\%) classes. There findings are also validated by the qualitative visualization of the results that is shown in Fig.~\ref{fig:no_dilation_compare}. We therefore consider that the low-level features encoder and merged dilated convolutions can obtain better local and global context information and thus segment both multi-scale objects (such as buildings and surface) and small classes (such as cars) accurately.

\subsection{Generalization}
Our models demonstrated very good generalization capabilities on both the Potsdam and Vaihingen dataset. As shown in Table~\ref{tab:rgbresults}, there are only -0.2\% and -1.1\% gaps in terms of mean F1-score between our local validation sets and the hold-out test sets of the Potsdam and Vaihingen, respectively. It is also worth noting that our model is the only one that works equally well on both Vaihingen IRRG dataset and Potsdam RGB dataset, which outperforms the DST\_2 \cite{Sherrah16} model with 3.9\% and 0.6\% higher F1-score on Vaihingen and Potsdam dataset, respectively, as shown in Tables~\ref{tab:potsdam_scores} and \ref{tab:vaihingen_scores}. Furthermore, our model achieves better performance (+0.5\%) in terms of mean F1-score with fewer labeled training data than FuseNet+OSM \cite{audebert2017joint} that used OpenStreetMap (OSM) as an additional data source.

\begin{figure}[htpb!]
 \centering
  \includegraphics[width=0.48\textwidth]{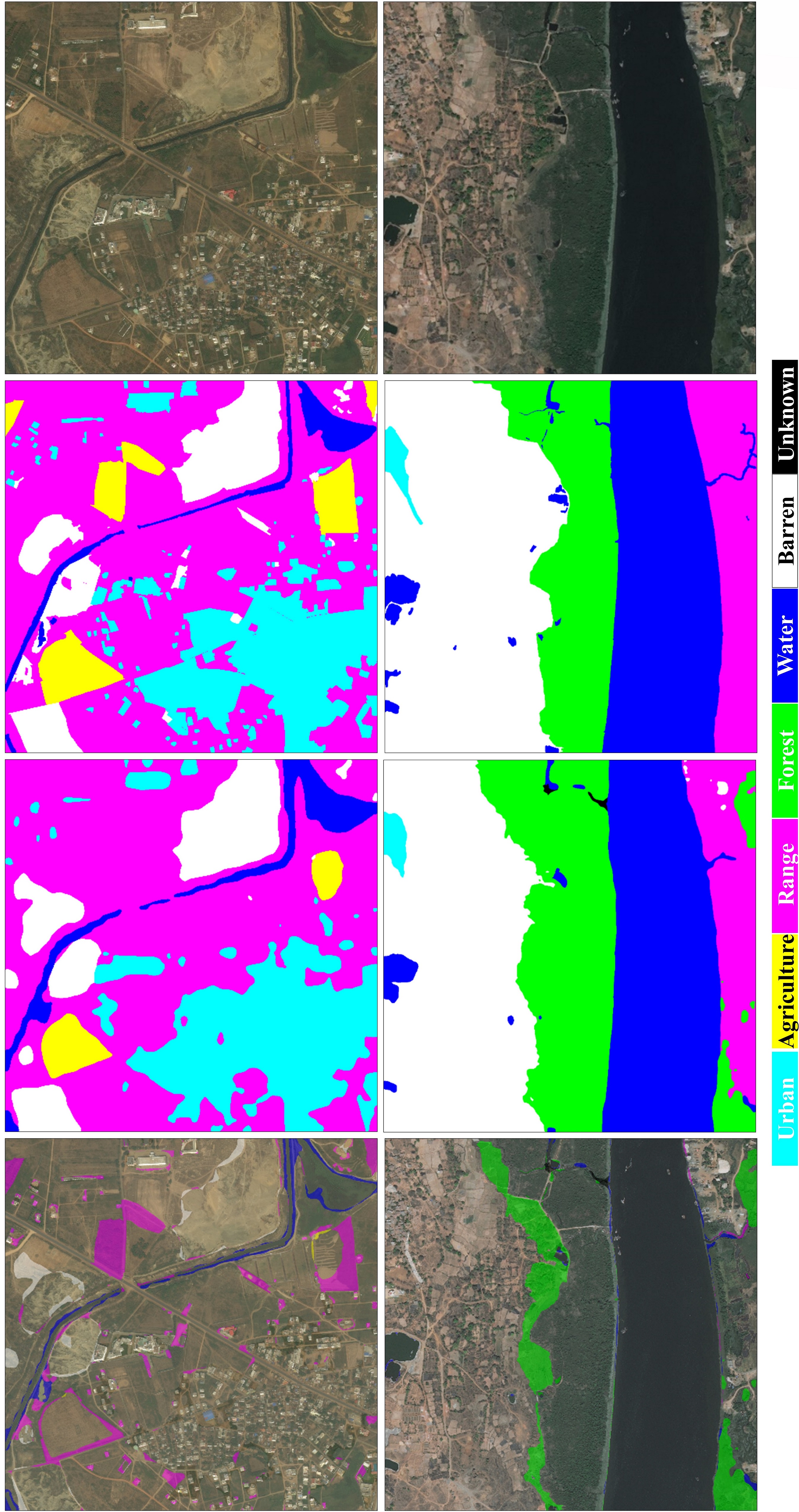} %,height=8.5cm]{testallpng.png}
  \caption{Mapping results on local validation images of DeepGlobe. From top to bottom, the input satellite images, the ground truths, the predictions of DDCM-SER50 model, and the errors of predictions.}
  \label{fig:db_test}
\end{figure}

However, we observed a bigger performance drop (approximately -15.9\%) on the DeepGlobe dataset when comparing results on the hold-out test set (mIoU: 56.2\%) and local validation results (K-Fold avg. mIoU: 72.1\%) as shown in Table~\ref{tab:kfold_results}. We see there is higher uncertainty (more false predictions) between range land (magenta), agriculture land (yellow), and forest (green) from the  confusion matrix between classes for our model in Table~\ref{tab:cm}. Note that the model incorrectly classified some agriculture land to rangeland, and predicted some rangeland as forest. These observations are also supported by the qualitative visualization of errors of our predictions as shown in Fig.~\ref{fig:db_test}. Furthermore, in our experiments on the DeepGlobe data, we found that there are some annotation inaccuracies, mainly introduced by highly ambiguous objects and lower ground resolutions. What is worse, we observed that the hold-out test images have different contrast and darker shadows than in the training set \cite{samy2018nu}. This obviously affected the model's final performance on the test sets. 
% \commentAB{What do you mean by "lower altitude"? Altitude is used for satellites and aerial vehicles. Since the DeepGlobe data is satellite data, the altitude is constant for all time. If you refer to location on ground, the correct word is "elevation".}

\begin{table}[htbp!]
  \centering
  \caption{IoU scores of our 5 K-fold models on local validation sets of DeepGlobe dataset.}
  \resizebox{\columnwidth}{!}{
    \begin{tabular}{c|cccccc|c} 
    \textbf{K-Fold} & \textbf{Urban} & \textbf{Agriculture} & \textbf{Range} & \textbf{Forest} & \textbf{Water} & \textbf{Barren} & \textbf{mIoU} \\
    \hline
    k0    & 0.783 & \textbf{0.901} & \textbf{0.488} & 0.760  & 0.605 & 0.723 & 0.710 \\
    k1    & 0.735 & 0.876 & 0.391 & 0.772  & 0.730 & \textbf{0.739} & 0.707 \\
    k2    & \textbf{0.821} & 0.873 & 0.381 & 0.757  & 0.832 & 0.723 & 0.731 \\
    k3    & 0.796 & 0.883 & 0.421 & 0.789  & 0.849 & 0.716 & \textbf{0.742} \\
    k4    & 0.724 & 0.830 & 0.431 & \textbf{0.795} & \textbf{0.857} & 0.654 & 0.715 \\
     \hline
    \textbf{Avg.} & \textbf{0.772} & \textbf{0.872} & \textbf{0.422} & \textbf{0.775} & \textbf{0.775} & \textbf{0.711} & \textbf{0.721} \\
    \hline
    \end{tabular}
}
  \label{tab:kfold_results}%
\end{table}%

 \begin{table}[htbp!]
 \centering
  \caption{Normalized confusion matrix on a local validation set of DeepGlobe data.}
  \includegraphics[width=0.98\columnwidth]{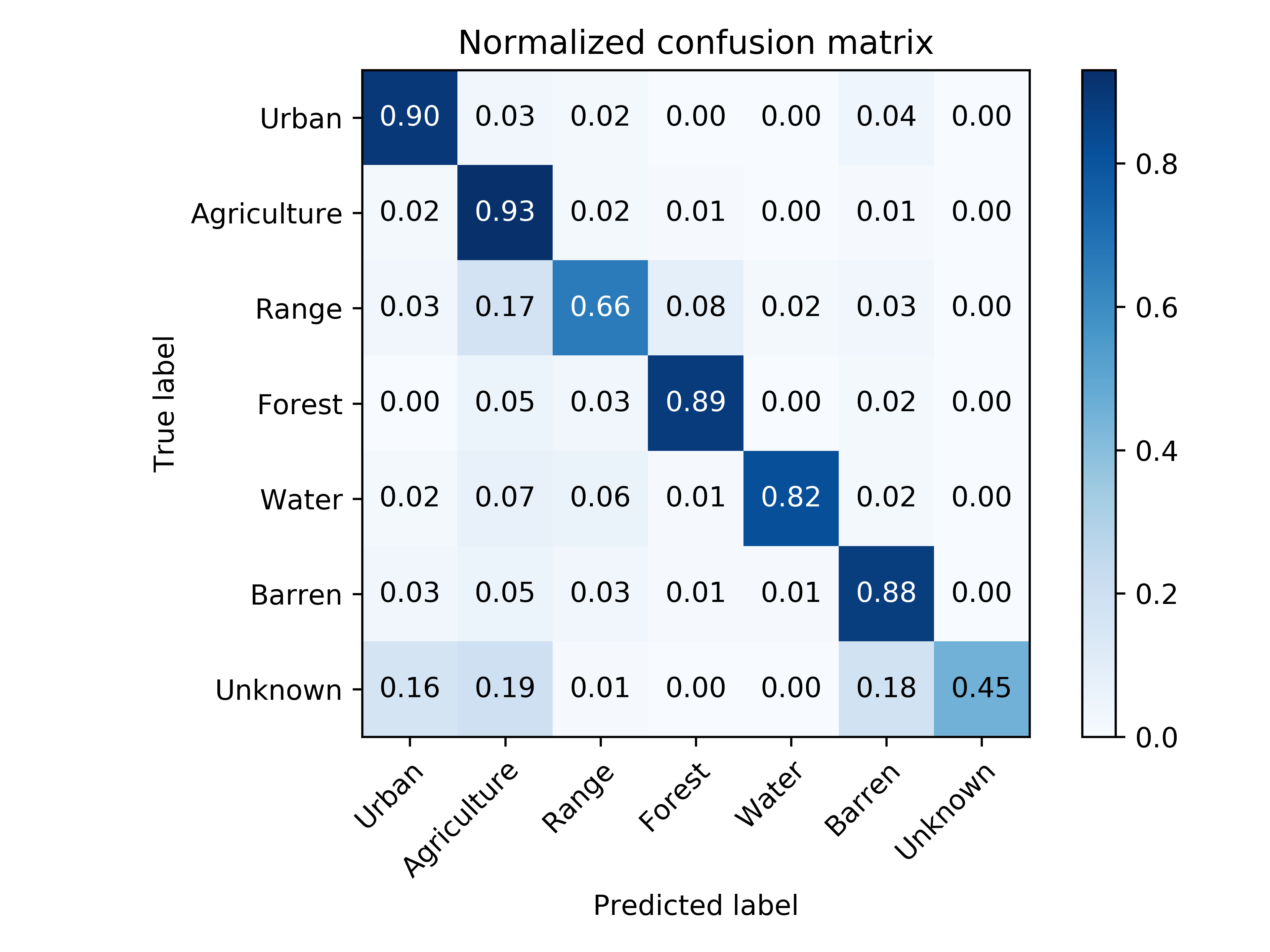} %,height=8.5cm]{testallpng.png}
  \label{tab:cm}
\end{table}

\section{Conclusions} \label{concl}
In this paper, we presented a dense dilated convolutions merging (DDCM) architecture for land cover classification for aerial imagery. 
The proposed architecture applies dilated convolutions to learn features at varying dilation rates, and merges the feature map of each layer with the feature maps from all previous layers. On both the Potsdam and Vahingen datasets, our single model based on the DDCM-Net architecture achieves the best mean F1-score compared to the other architectures, but with much fewer parameters and feature maps. DDCM-Net is easy to adapt to address a wide range of different problems, is fast to train, and achieves accurate results even on small datasets. The variants of our DDCM-Net by using different combinations of dilations and densities for the DeepGlobe dataset also demonstrated better performance, but consumed much fewer computation resources compared to other published methods.
% if have a single appendix:
%\appendix[Proof of the Zonklar Equations]
% or
%\appendix  % for no appendix heading
% do not use \section anymore after \appendix, only \section*
% is possibly needed

% use appendices with more than one appendix
% then use \section to start each appendix
% you must declare a \section before using any
% \subsection or using \label (\appendices by itself
% starts a section numbered zero.)
%

% \appendices
% \section{Proof of the First Zonklar Equation}
% Appendix one text goes here.

% % you can choose not to have a title for an appendix
% % if you want by leaving the argument blank
% \section{}
% Appendix two text goes here.

% use section* for acknowledgment
\section*{Acknowledgment}
This  work  is  supported  by  the  foundation  of  the  ResearchCouncil of Norway under Grant 220832. The authors would also like to thank the ISPRS for making the Potsdam and Vaihingen datasets publicly available. 

% The authors would like to thank...

% Can use something like this to put references on a page
% by themselves when using endfloat and the captionsoff option.
\ifCLASSOPTIONcaptionsoff
  \newpage
\fi

% trigger a \newpage just before the given reference
% number - used to balance the columns on the last page
% adjust value as needed - may need to be readjusted if
% the document is modified later
%\IEEEtriggeratref{8}
% The "triggered" command can be changed if desired:
%\IEEEtriggercmd{\enlargethispage{-5in}}

% references section

% can use a bibliography generated by BibTeX as a .bbl file
% BibTeX documentation can be easily obtained at:
% http://mirror.ctan.org/biblio/bibtex/contrib/doc/
% The IEEEtran BibTeX style support page is at:
% http://www.michaelshell.org/tex/ieeetran/bibtex/
%\bibliographystyle{IEEEtran}
% argument is your BibTeX string definitions and bibliography database(s)
%\bibliography{IEEEabrv,../bib/paper}
%
% <OR> manually copy in the resultant .bbl file
% set second argument of \begin to the number of references
% (used to reserve space for the reference number labels box)

\bibliographystyle{IEEEtran}
\bibliography{bibtex/my.bib}

% \begin{thebibliography}{1}

% \bibitem{IEEEhowto:kopka}
% H.~Kopka and P.~W. Daly, \emph{A Guide to \LaTeX}, 3rd~ed.\hskip 1em plus
%   0.5em minus 0.4em\relax Harlow, England: Addison-Wesley, 1999.

% \end{thebibliography}

% biography section
% 
% If you have an EPS/PDF photo (graphicx package needed) extra braces are
% needed around the contents of the optional argument to biography to prevent
% the LaTeX parser from getting confused when it sees the complicated
% \includegraphics command within an optional argument. (You could create
% your own custom macro containing the \includegraphics command to make things
% simpler here.)
%\begin{IEEEbiography}[{\includegraphics[width=1in,height=1.25in,clip,keepaspectratio]{mshell}}]{Michael Shell}
% or if you just want to reserve a space for a photo:

% insert where needed to balance the two columns on the last page with
% biographies
%\newpage

% You can push biographies down or up by placing
% a \vfill before or after them. The appropriate
% use of \vfill depends on what kind of text is
% on the last page and whether or not the columns
% are being equalized.

%\vfill

% Can be used to pull up biographies so that the bottom of the last one
% is flush with the other column.
%\enlargethispage{-5in}

% that's all folks
\end{document}